\documentclass[twocolumn,10pt]{asme2ej}

\usepackage{epsfig} 
\usepackage{amsmath} 
\usepackage{comment}

\usepackage{hyperref}
\hypersetup{%
  breaklinks,
  bookmarksnumbered=true,
  bookmarksopen=true,
  colorlinks=true,
  linkcolor=black,
  urlcolor=black,
  citecolor=black
}

\title{Developing a Simple Model for Sand-Tool Interaction and Autonomously Shaping Sand}

\author{Wooshik Kim, Catherine Pavlov, and Aaron M. Johnson
    \affiliation{
	Robomechanics Laboratory\\
	Mechanical Engineering\\
	Carnegie Mellon University\\
	Pittsburgh, Pennsylvania\\
    Email: brikim13@gmail.com, \{cpavlov,amj1\}@andrew.cmu.edu
    }	
}

\begin{document}

\maketitle    

\begin{abstract}
{\it 
Autonomy for robots interacting with sand will enable a wide range of beneficial behaviors, from earth moving for construction and farming vehicles to navigating rough terrain for Mars rovers. The goal of this work is to shape sand into desired forms. Unlike other common autonomous tasks of achieving desired state of a robot, achieving a desired shape of a continuously deformable environment like sand is a much more challenging task. The state of robot can be described with a couple of states--x, y, z, roll, pitch, yaw--but the desired shape of sand can not be described with just a few values. Sand is an aggregation of billions of small particles. After simplifying the model of sand and tool interaction by looking only at the surface of the heightmap, we can formulate the problems into something that is still high dimensional (hundreds to thousands of state dimensions) but much more solvable. We show how this problem can be formulated into a graph search problem and solve it with the A-star algorithm and report preliminary results on using deep reinforcement learning methods like Deep Q-Network and Deep Deterministic Policy Gradient.

}
\end{abstract}




\begin{figure}[h!]
\centering
\includegraphics[width=0.48\textwidth,trim=10 0 0 10,clip]{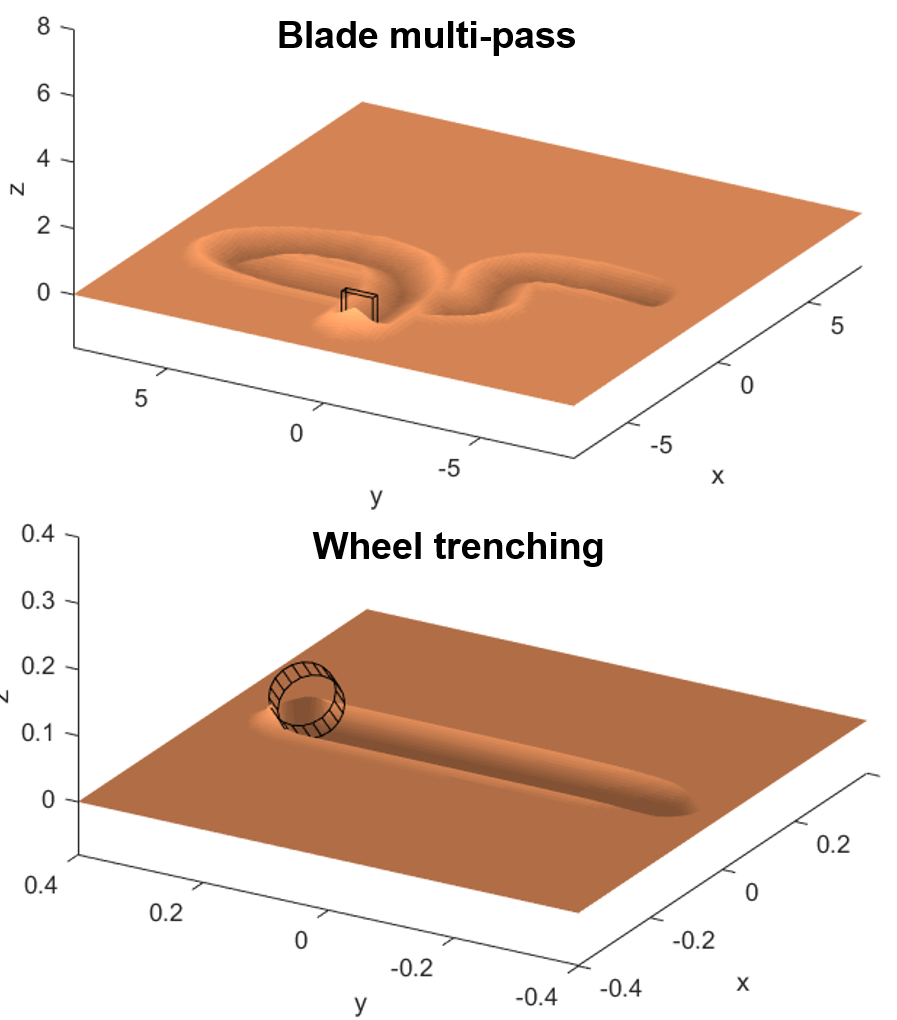}
\caption{Blade multipass example and wheel trenching example}
\label{fig:multi_wheel}
\end{figure}

\section{Introduction}
In this work, we introduce an interesting new problem: shaping a continuously deformable environment, sand. Shaping sand is a challenging task that can benefit both the industry and the academy. In construction, dumping operations (removing soil) account for 25-50\% of total operational costs \cite{earthdumping}. In the US alone, gross output of construction and agriculture industry accounts for \$1,970 billion \cite{gdpreport}. The global market for earth moving equipment valued at \$81.5 billion in 2017 and expected to cross \$145 billion by 2026 \cite{earthmovingreport}. Many big industries like John Deere, Caterpillar, Komatsu, and CNH are already investing on autonomous construction vehicles, which could benefit from autonomous operations. 

Autonomous sand shaping also has potential applications in planetary exploration; enabling rovers to modify terrain can greatly increase their capabilities \cite{Pavlov}. For example, by digging a trench a rover can access subsurface soil for sampling; by pushing sand into a ramp a rover could climb otherwise untraversable terrain; or by levelling uneven terrain it could construct a landing site for future missions. In this context, sand would be shaped either by a simple, lightweight tool or the existing mobility systems of the robot (e.g.\ wheels), as planetary missions are highly weight constrained. 

In this paper we categorize different types of sand modelling methods and discuss works on automation of earthmoving. We then provide our own solution to shaping sand along with the necessary models. This paper is organized as follows: First, in Section~\ref{sec:previous}, we will look at four different approaches of simulating sand and robot interaction. This survey will reveal different approaches that we integrate into our model. We also look at how other researchers have approached the problem of shaping sand autonomously in Section~\ref{sec:earthmoving}. Our method for simulating sand--robot interaction is presented in Section~\ref{section:sand_simulation} followed by a comparison of a trench profiles with other model and sand experiment. Then algorithms for planning motions to shape the sand are presented in Section~\ref{sec:planning}, using A*. Finally, we will share our insights and analysis in Section~\ref{sec:analysis} and conclusions in Section~\ref{sec:conclusion}.

\section{Previous Works}
\label{sec:previous}
Numerous models and approaches have been proposed by researchers from different fields of studies to describe the dynamics of sand. Terramechanics research describes the dynamic effects of sand with respect to vehicles, and it also studies how microscopic interactions among grains result in sand's macroscopic behavior like soil compression and erosion \cite{bekker, wong}. Computer graphics research mostly focuses on how to render realistic animations \cite{cundall}. In robotics, researchers focus on implementing models that are computationally efficient yet accurate in their steady state behavior to use for control and planning purposes \cite{scm2008, scm2016}. Here, we categorize these into four different general methods used for describing sand. 

\subsection{Solid Mechanics and Empirical Formulas}
The most extensive and iconic work in the mechanics of sand is that of Bekker \cite{bekker} and Wong \cite{wong}. Most of the analyses are based on extensive empirical results and application of theory of plastic equilibrium to the mechanics of vehicle-terrain interaction. Because of its accuracy and computational efficiency, Bekker's terramechanics are often implemented in conjunction with other simulator environments \cite{scm2008,pla-castells,scm2016}. It is also often used as a criteria for comparing accuracy of a simulation \cite{DEMpush}. Some of the most common results from terramechanics that finds their way to vehicle-soil simulators are: i) pressure sinkage relationship \cite{bekker,wong} to simulate vehicle wheel compressing soil; ii) maximum shear stress to simulate vehicle wheel slippage \cite{wong}; iii) force calculation for soil-tool interaction \cite{perumpral}. Force calculation proves quite useful in tillage and bulldozing applications and hence are often implemented in conjunction with other simulations \cite{pla-castells}.

\subsection{Discrete Element Methods}
The dynamics of sand is the result of many, many small grains interacting with each other. Therefore, the most accurate way to model it is with numerical simulations based on the discrete element method (DEM) proposed by \cite{cundall}. This approach studies granular materials by solving Newton's equation of motion for individual particles. The simulation of the dynamics of large collections of solid particles have become efficient enough to allow both the macroscopic description of the behavior and some microscopic details about local phenomena \cite{DEMpush}. The interaction between particles is defined by translational and rotational contact dynamics at their contact points. The contact force acting on a particle is decomposed into normal and tangential components where each component is modeled by using spring, dashpot, and a frictional slider. The number of particles used in DEM simulation is usually in the range of few hundred thousands \cite{DEMpush}.   

DEM methods are also often used in conjunction with other modeling methods to simulate soil-tool interaction. Holz generates particles locally where interaction takes place and uses Height Map (Section~\ref{section:height-map_approach} to represent the rest of the terrain \cite{demhybrid}. The addition of cohesive forces between particles (wet sand) enables simulation of cohesive soil as well \cite{DEMcohesive}. Although this method captures how the grains of sand behave in various situations, the computational load is heavy and therefore not appropriate for learning or planning. 

\subsection{Fluid Mechanics}
To overcome the computational load from discrete element methods and render realistic animations, researchers have borrowed ideas from fluid dynamics. Here Eulerian approaches are utilized rather than Lagrangian methods as DEM does. The material point method
keeps track of the force and velocity field at each grid
instead of keeping track of individual particles  \cite{materialpointmethod}. State of the art methods are able to render realistic animations of sand and fluid mixture in DreamWorks' most recent work \cite{sandFluidTampubolon}. These methods, although aesthetically beautiful, are not fit for robotic applications. Although it is less computationally expensive than discrete element methods, it is still computationally burdensome for robotic applications. Second, it inevitably comes with volume error because it does not track particles, which is an important aspect for shaping sand. Finally, it cannot provide a good way to analyze force interaction between the sand and the robot. 

\subsection{Height-map Approach} \label{section:height-map_approach}
To describe sand realistically, yet computationally efficiently for robotic applications, the height-map approach works the best. Instead of describing the entire volume of sand, as the two earlier approaches did, height-map approach considers the dynamics at the surface: by describing the surface with a height-map. Soil erosion, wheel interaction, and tool interactions (bulldozer and excavators) are interactions that happens on the surface of sand. Therefore with this simplification, this method can accurately model sand and tool interactions incredibly efficiently. Starting from the simple idea that sand piles up only up to certain angle, the angle of repose, one can model sand by relaxing the local slope of sand by ``flowing'' down the slope until it reaches the angle of repose. This phenomenological approach intrigued physicists in the 1990s starting from \cite{BTW}. Those studies were focused on the ``self-organized criticality'', the scale and time-invariant dynamics of soil erosion \cite{hwakardar,criticality2,criticality3}. These early studies focused on how this local governing behavior propagates to create an ``avalanche''.

Application of this height-map idea were developed further by understanding that the surface of sand behaves differently than the pile beneath. When there is a soil erosion, the soil on top rolls down the slope, rather than the pile beneath. Developing these ideas were Hwa and Kardar's anistropic ``Driven Diffusion Equation'' \cite{hwakardar} and Bouchad, Cates, Prakash, and Edwards BCRE method \cite{BCRE}. 

Actually using these ideas to simulate and visualize interactions were done recently. For simulating excavator and sand interaction, theory of BCRE was applied to simulate the time evolution of sand pile \cite{pla-castells}. This work also used an additional height-map to describe sand on the flat excavator tool. When sand fell out of bounds from the excavator, that sand was added to the corresponding location on the height-map of ground. Another interesting take on using height-map approach is from \cite{hsmap}. Describing each collision object in the simulator with \textit{Height Span Map} (HS Map), sand on top of each HS map follow same rule for soil erosion. So sand can sit inside a concave bucket or on top of its convex outer surface. Simulator by Rainer Krenn et al. from Germany Aerospace Center (DLR) combines the height-map approach and Bekker's terramechanics \cite{scm2008, scm2016}. This work merged the ideas from the height-map approach with the multi-body physics engine SIMPACK. In addition to soil erosion with height-map approach, this work incorporates soil-compaction and wheel slip from Bekker's terramechanics \cite{theoreticalsoilmechanics}. Height of soil is reduced appropriately on the trail where rover passed by according to soil-compaction equation. Wheel slip is calculated beforehand according to Janosi-Hanamoto's equation for the rover. This paper's approach is most similar to this work as Krenn also proposed the model for robotic application. However the focus of this work is to shape the sand, meaning we are more interested in how the sand looks after interaction.

DEM and fluid mechanics methods are very successful for rendering sand dynamics, but they are computationally too expensive for robot application and is actually unnecessary. For most robotic application, like bulldozer and excavator, robots move slow relative to the flow of sand. So how the sand flows down the slope over time is not of concern, the resulting surface profile is. Therefore, the empirical formulas from Bekker's terrachmechanics comes in handy. But for simulating sand, the height-map method is computationally the most efficient, accurate, and easy for describing sand's steady state behavior. Therefore, in this paper the height-map method is used.

\subsection{Automation in Earthmoving}
\label{sec:earthmoving}
Most of the automation research in earthmoving has been focused on dig and unload tasks for excavators \cite{singh_art}. Many interesting approaches have been proposed to aid workers in the construction industry from tele-operation of construction vehicles \cite{teleoperation_starzewski} to planning sequences of actions to produce an overall desired shape \cite{romero_planner}. Romero creates a tree of excavation sections and actions to generate a plan that minimizes task completion time. Another interesting take is Sanjiv Singh's planner for robotic excavators where the agent determines the optimal dig parameters (angle of attack, length of dig, and height to enter) for given constraints (reachability, volume, shaping, and force) per action \cite{singh_planner}. More recently, there has been couple of approaches on creating an autonomous agent for bulldozers. Hirayama developed an agent for bulldozer dumping operation by segmenting mound into blocks and using integer linear programming to find optimal sequence of trajectories that will dump the material over the edge position \cite{bulldozerplanning2018, bulldozerplanning2019}. 

The complexity of soil to robot interaction makes it challenging to develop a decent controller for both low level tracking and high level planners \cite{challenges}. Recent development in artificial intelligence and machine learning may hold the key for tackling the high dimensionality of sand-robot interaction. For example, Clarke developed a neural network that uses audio feedback to scoop the right amount of granular materials \cite{clarke_scoop}. 

\section{Method: Simple Simulator for Sand and Robot Interaction} \label{section:sand_simulation}

The simulation of sand-robot interaction happens in two steps. First an \textit{action} is taken. An \textit{action} is when a robot changes state, either interacting with sand or not. For bulldozing tasks, as we consider in this work, the \textit{action} is pushing sand. For excavator, it would be removing (scooping) and depositing sand. For mobile robots, it would be leaving track behind, which is similar to pushing sand. After each \textit{action}, an \textit{update} step iteratively displaces sand to simulate soil erosion. The number of steps it takes to reach steady state depends on the resolution; a finer sand profile requires more iterations. For the planning examples in Section~\ref{sec:planning}, the size of the height map is 32 by 32 and it takes around 50 steps to reach steady state. For the trench profile comparison in Section~\ref{sec:profilecomp}, the size of height map is 160 by 160 and it takes around 140 steps to reach steady state. 

This simulator does not consider the transient response of the soil erosion and the dynamics of robot are assumed to be much slower compared to the sand. Also the dynamics between the robot and sand is also neglected, there is no force feedback to the robot (i.e. we assume the robot is stiff). The focus of this work is to find trajectory of the robot that produce a certain height map. 

\begin{figure}[h!]
\centering
\includegraphics[width=0.48\textwidth]{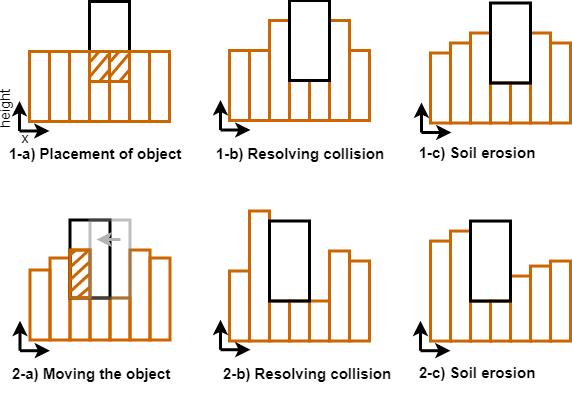}
\caption{Tool interaction and soil erosion during \textit{update} after two \textit{actions}: placement of object and movement of object}
\label{fig:soiltool}
\end{figure}

\subsection{Soil Erosion}
The proposed model represents the sand's surface with a height-map as described in Section \ref{section:height-map_approach}. The volume of granular material is divided into vertical columns in an equally divided grid. Height values are saved in a two dimensional matrix. There are two main components of this simple model: soil erosion and tool interaction. Soil erosion makes sure the profile of sand satisfies the angle of repose constraint. Tool interactions allows a robot (robot arm, bulldozer, or rover) to push and move sand around. 

Soil erosion starts from the simple idea that sand flows downward if it is piled too steeply. 
The maximum slope sand can reach before grains start rolling is called the angle of repose, and this is a property of sand that changes depending on things like grain size and humidity. 
After each \textit{action}, the local slope of the sand profile is evaluated at every point of the grid and compared to the angle of repose during each \textit{update}. The amount of sand, and consequently the height of each grid, that flows during steps of the \textit{update} is as follows:   

\begin{align}
q_{(i,j)\leftarrow(k,l)} &=
\label{eq:erosion}
\begin{cases}
k \Delta x(\frac{\partial h}{\partial x} + b_{repose}), & \frac{\partial h}{\partial x} < - b_{repose}  \\
k \Delta x(\frac{\partial h}{\partial x} - b_{repose}), & \frac{\partial h}{\partial x}>b_{repose}  \\
0, & \text{otherwise} 
\end{cases} \\
\Delta h_{i,j} &= \frac{1}{A} \sum_{(k,l)} q_{(i,j)\leftarrow(k,l)}
\end{align}

At each step of the \textit{update}, each height of the height-map matrix is updated according to the local slope with its neighbors (four connectivity or eight connectivity). The number of steps needed to reach steady state is proportional to the square of grid resolution (complexity of $N^2$). Then the question that remains is how much of the soil imbalance flows during each step: the value of $k$. The $k$ must be chosen to allow maximum flow during each step without overshooting. For eight connectivity $k=\Delta x^2/8$ worked best. 

\subsection{Tool Interaction} 
Each object is represented by collision points at each grid point (step \textit{a} of Figure \ref{fig:soiltool}). For tool interaction, the overlapping volume of the height-map that is in collision with the tool is moved to an adjacent grid (step \textit{b}) in the direction of the tool movement. At initial contact, the sand is moved radially outwards from the center of the tool (step \textit{1-b}). After overlapping volumes are moved to adjacent grid cells, soil erosion iteratively revises the local slope by updating the profile (step \textit{c}). Soil does not flow into the grid occupied by the tool (though instead one could represent tools as height span map as proposed by \cite{hsmap}). 

\subsection{Results: Trench Profile Comparison with Experiment and Soil Model}
\label{sec:profilecomp}

The sand model presented in this section was validated with a set of experiments originally presented in \cite{Pavlov} in the soft-soil testbed in Carnegie Mellon's Field Robotics Center. These experiments collected 2D profiles of trenches of wheels towed across a prepared sand surface at fixed sinkage ($h_0$) and angle ($\beta$) at a velocity of 3 cm/s. Each trench was then imaged with a LIDAR scanner at 12 locations along with length to collect a surface profile with 300-400 data points. The LIDAR scanner was found to have a standard deviation of 0.6mm error for measurements of a flat sand surface, which is considerably smaller than the typical feature scale in these experiments. The wheel measurements and soil parameters for these experiments are listed in Table \ref{table:trench_params}.

The sand model presented here was compared both to the trench scans and to the theoretical model described in \cite{Pavlov}, with results summarized in Table \ref{table:results_grouser}. Compared to LIDAR scans of the trench surface, the sand simulator had an average error of 2.2 mm and median error of 1.7 mm along the trench profile, with a 0.9mm error in the predicted depth of the trench. Compared to \cite{Pavlov}, the sand simulator had an average error of 1.1 mm and median error of 0.4 mm, with a depth error of 0.6mm. Given the accuracy of the LIDAR scanner, the soil simulator was able to predict the shape of the trench very well. The close match to the trenches predicted by \cite{Pavlov} is notable given the different approaches taken to modeling soil flow.

\begin{figure*}[t] \label{fig:profile_comparison}
  \includegraphics[width=\textwidth]{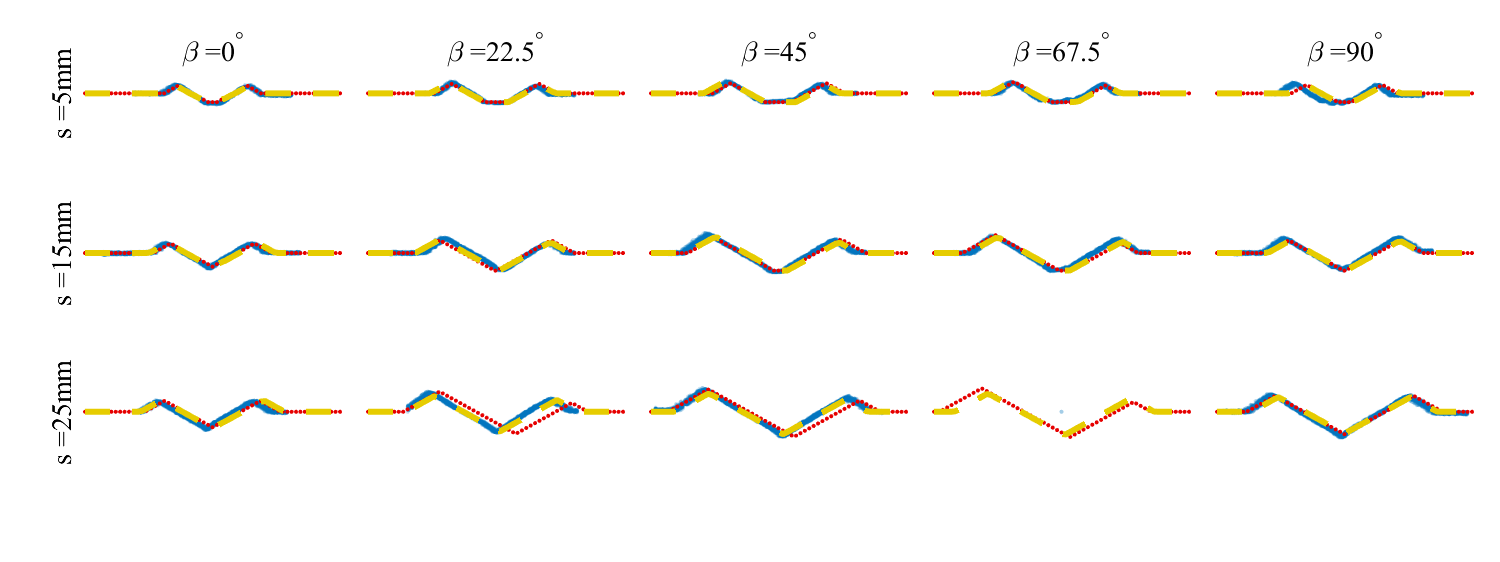}
  \caption{Plots of measured (blue), trench model predicted (red) \cite{Pavlov}, and simulation (yellow) trenches for varied slip angle and sinkage. All 12 measurements of each trench (blue) are shown. Note that there is no trench measurement for the 25mm sinkage trial at $\beta = 67.5^\circ$ due to excessive deflection of the test rig. }
\end{figure*}

\section{Method: Planning Algorithms}
\label{sec:planning}
The goal of this work is to manipulate sand into arbitrary shapes autonomously. What kinds of arbitrary shapes are there? First, within the construction industry, the tasks can be divided according to the types of construction vehicles they require: 1) digging and excavating with excavators 2) leveling and trenching with bulldozers, and 3) building a mound and ramp with excavators or bulldozers. In this work, we will only consider trenching tasks with bulldozers. The goal is to make a certain shape trench after a sequence of pushes from an initial state of flat surface. We imagine a scenario where a robotic arm that has a flat blade for its end-effector pushes the sand. The robot will always push sand perpendicularly, so the goal is to find a trajectory sequence that will shape sand into a desired shape. We also assume desired trenches are at constant depth. 

\begin{figure}[t]
\centering
\includegraphics[width=0.48\textwidth]{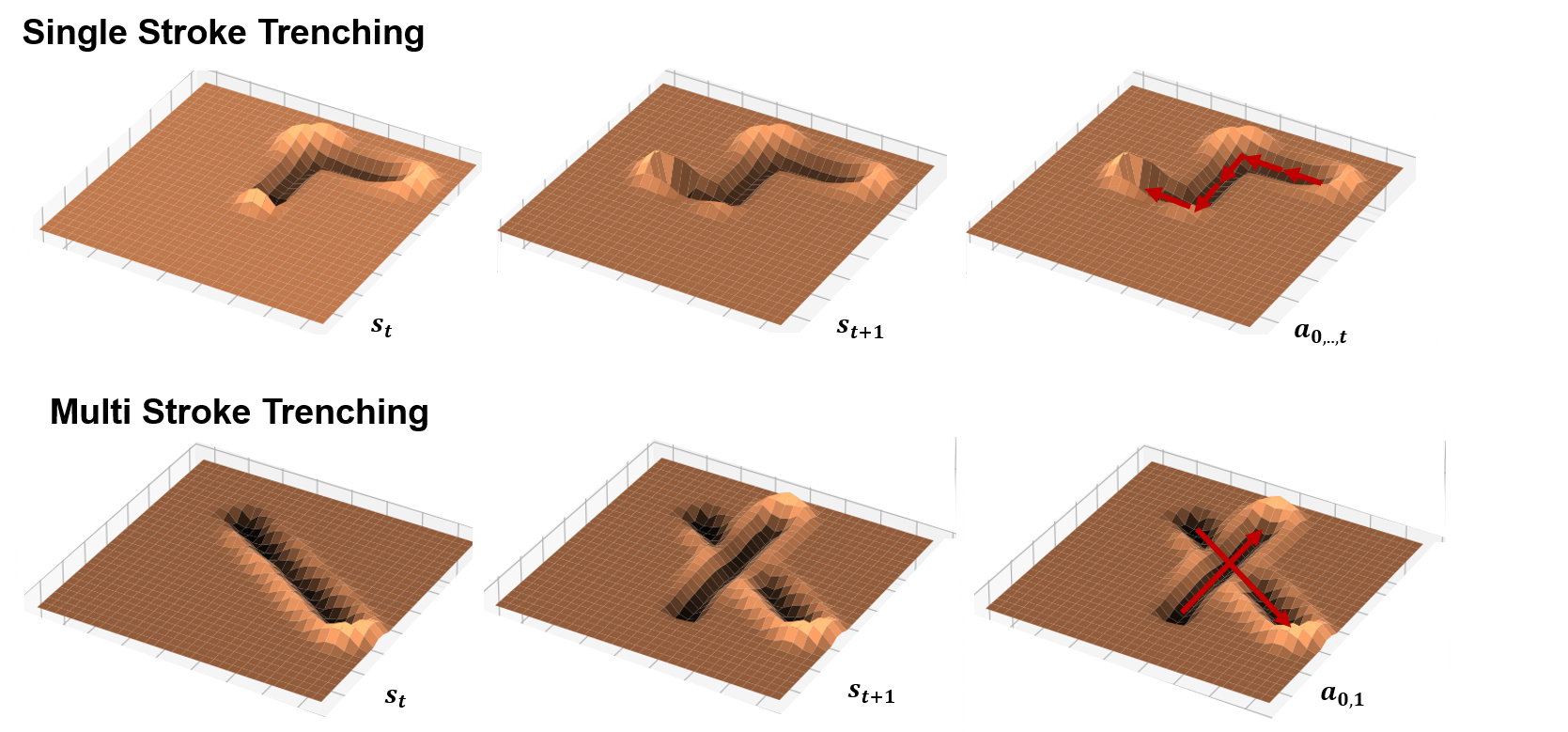}
\caption{Two Environment Settings: single stroke trenching and multi stroke trenching} 
\label{fig:single_multi}
\end{figure}

\subsection{Problem Definition} 
Given a goal height-map $H_g$ and a initial level surface $H_0$, the objective is to find a sequence of actions to minimize the difference between $H_g$ and $H_n$, where $n$ is the final time. The state at time $t$ is $s_t = (H_t, X_t)$, where $X_t$ describes the state of the robot. 

We then formulate two problem scenarios for the trenching task as shown in Figure \ref{fig:single_multi}. The \textbf{single stroke shaping} is where the end-effector never leaves the trenching depth and has to dig the trench in one go. At each step the agent will choose one of the four discrete actions, $a_t \in (north, east, south, west)$. This formulation allows discrete planners like A-star (A*) and Deep Q-Network (DQN) \cite{a-star, dqn}. The \textbf{multi-stroke shaping} assumes a more relaxed scenario where at each step, the agent chooses a start point and end point to perform a variable length of a single straight push. Here, the action is continuous $a_t = (x_0,y_0,x_1,y_1)$, where $x_0,x_1 \in [x_{min},x_{max}], y_0,y_1 \in [y_{min},y_{max}]$. This formulation will require continuous planners like Deep Deterministic Policy Gradient (DDPG) \cite{ddpg}.

\subsection{A-star Planner for Trenching}
\begin{figure}[t]
\centering
\includegraphics[width=0.48\textwidth]{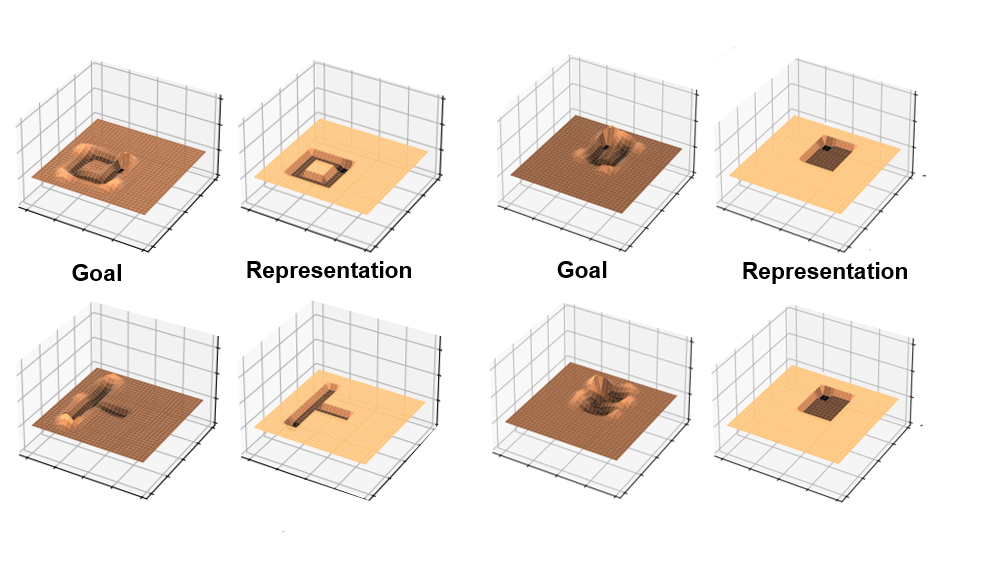}
\caption{Simple representation of sand height-map for A-star planning}
\label{fig:representation}
\end{figure}

For \textbf{single stroke shaping}, we can simplify the sand state into a binary map. As shown in Figure \ref{fig:representation}, we represent the continuous heightmap with a binary height map of ones and zeros. Each push ``digs'' a portion of the map into zeros. But this simplification entails some loss of expressivitiy, even with discrete actions and fixed depth. For example, the box in Fig 5 right has two maps that result in the same representation. Nevertheless, this assumption is still acceptable because we are 1) assuming a constant blade depth for pushing the sand, 2) interested only in the dug portion of the sand and not interested in the pile that accumulates adjacent to the trench. This reduces the dimension of the problem further so we can use graph-search algorithm without continuously expanding the search tree. Yet it is relevant and accurate enough to dig trenches into desired shape. 

The A-star algorithm, which is a heuristic optimal weighted sequential planner, uses an estimate of the cost-to-go $\hat{h}(x)$, in addition to cost-so-far $g(x)$, to expand the search tree efficiently \cite{a-star}. The objective here is to reach desired height-map while taking the minimum number of \textit{actions}. So the cost should include the height-map difference and the distance traveled. We combine the two costs to estimate the optimal path cost $f(x)=g(x)+\hat{h}(x)$. The cost-so-far is the distance traveled $g(x)=\sum \Delta s$, and the distance-to-go is $\hat{h}(x) = \alpha \sum \mid H_g - H_t \mid$, where the weight $\alpha$ is determined so that the value of sand displaced after an \textit{action} is the same as the step distance, $\Delta s = \alpha \Delta H$. Our approximation is admissible as $\hat{h}(x) < h(x)$ since we can not achieve the goal shape with less than $\Delta H$ left.

\subsection{Results: A-star Planner Trenching in a Single Stroke}
\begin{figure}[h!]
\centering
\includegraphics[width=0.48\textwidth]{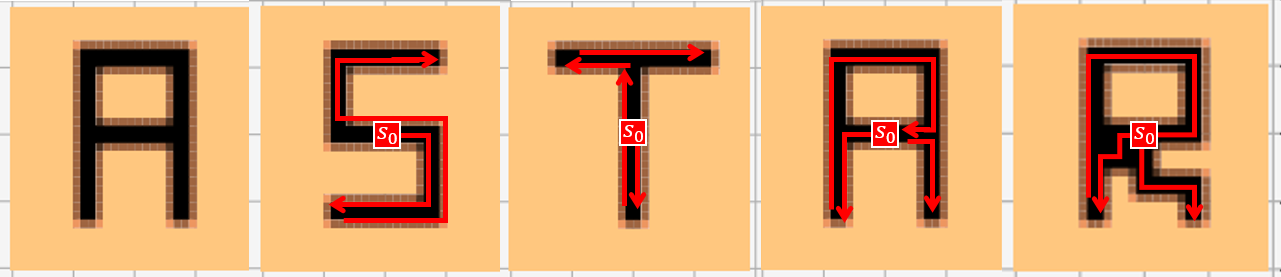}
\caption{Optimal Path for Shaping with Single Stroke}
\label{fig:astar}
\end{figure}

Once we formulate the problem into a graph search algorithm, we can use A* for \textbf{single stroke shaping}. Figure~\ref{fig:astar} shows a few shapes the algorithm achieved with actions overlaid. All planned path are optimal; it achieved the desired height map in the minimum number of steps.

After discretization of the state space into the binary heightmap, the problem of \textbf{single stroke shaping} becomes more feasible to solve. However, the number of nodes the algorithm has to open to find the optimal path quickly explodes as soon as it has to do multiple passes (i.e. come back the way we came). For each of the examples in Figure~\ref{fig:astar}, the number of nodes opened and steps it took were: A-18,532 nodes and 23 steps; S-788,492 nodes and 27 steps; T-23,864 nodes and 18 steps; R-28,096 nodes and 25 steps (though note that the state of the map can take any of $2^49$ possible values for a relatively small 7x7 grid, so these are still a small fraction of the total state space).

\subsection{Deep Reinforcement Learning for Shaping Sand}

Graph search algorithms like A-star quickly becomes infeasible for tasks involving continuous action and non-trivial shapes. We will apply deep reinforcement learning to achieve a desired height map.
Deep reinforcement learning (RL) using convolutional neural network proved to be useful for tackling problems with large state space (i.e. input is a three channel 2-D image). Our task can utilize such tools as the heightmap is like a 2-D image. In this work, we will explore two popular RL algorithms: Deep Q-Network (DQN) and Deep Deterministic Policy Gradient (DDPG) \cite{dqn, ddpg}. 

\textbf{State and Transition Function} The state space is constructed by all possible information that the agent can observe in the environment. We define a state with two parts: the goal height-map and current height-map, $s_t = (H_g, H_t)$. The transition function $s_{t+1}=trans(s_t,a_t)$ gives the transition process between states, which is implemented by the sand simulator developed above. 

\textbf{Action} The action space is the set of actions that the agent can perform. An action $a_t$ is a set of parameters that control the position of our robot's end-effector at step $t$. We define the behavior of an agent as a policy function $\pi$ that maps states to deterministic actions, $\pi:S \rightarrow A$. At step $t$, the agent observes state $s_t$ then gives the stroke parameters of the next stroke $a_t$. The state evolves based on the transition function $s_{t+1}=trans(s_t,a_t)$ until going on for $n$ steps. For DQN, we use \textbf{single stroke shaping} scheme where the actions are discrete. For DDPG, we use the \textbf{multi-stroke shaping} scheme where the actions are continuous. The points determine the start and end points of one stroke. 

\textbf{Reward} The reward function acts to evaluate the action decided by the policy. Selecting a suitable metric to measure the difference between the goal height-map and current height-map is crucial for the trenching agent. The reward is designed as, 
\begin{align} \label{eq:reward}
r(s_t,a_t)=L_t-L_{t+1}
\end{align}
where $r(s_t,a_t)$ is the reward at step $t$, $L_t$ is the measured loss between $H_g$ and $H_t$ and $L_{t+1}$ is the measured loss between $H_g$ and $H_{t+1}$. We used $l_2$ distance for the loss function in this work. 

To make sure the final height-map resembles the goal height-map, the agent should be driven to maximize the cumulative rewards in the whole episode. At each step, the objective of the agent is to maximize the sum of discounted future reward $R_t=\sum^T_{i=t} \gamma^{(i-t)}r(s_i,a_i)$ with a discounting factor $\gamma \in [0,1]$.

\subsection{Results: Deep RL for trenching alphabets}

\begin{figure}[h!]
\centering
\includegraphics[width=0.48\textwidth]{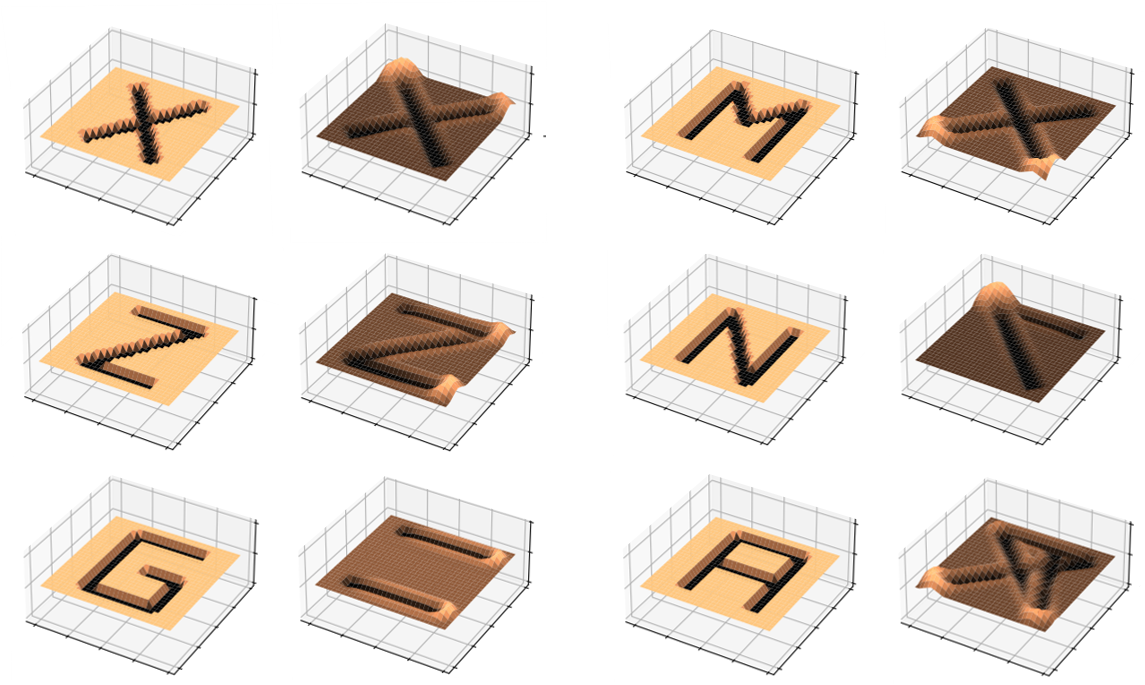}
\caption{Trenching Alphabets with DDPG}
\label{fig:alphabets}
\end{figure}

To test our algorithm we set the goal shape to be trenched alphabets. The algorithm we are currently testing is Deep Deterministic Policy Gradient (DDPG) with hindsight experience replay (HER) \cite{her}. With the hindsight experience replay, we add additional experience with the new goal set as the next state after each action. This speeds up training in the sparse reward environment. Figure~\ref{fig:alphabets} shows some example results after 10,000 episodes. So far the best results typically have two correct letters (X and Z) and 20/26 letters with at least one correct stroke. Further improvement in the network structure and additional training will be necessary to achieve a practical level of success.

\section{Analysis} 
\label{sec:analysis}
\subsection{Simulation of Sand Robot Interaction}
\begin{figure}[h!]
\centering
\includegraphics[width=0.48\textwidth]{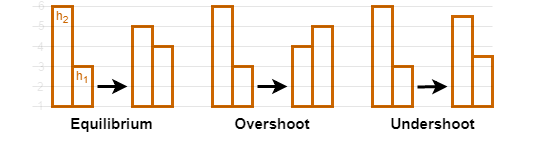}
\caption{The flow rate should be just right to reach equilibrium in the least number of steps. If too much is transferred to the adjacent block, the local slope changes sign or becomes zero. If not enough is transferred, it will take more steps to reach equilibrium.}
\label{fig:right_k}
\end{figure}

There were some insights we learned while trying to speed up the simulation. First is choosing the flow rate. To find the optimal $k$, the flow rate from Equation \ref{eq:erosion}, consider two height blocks that are not in equilibrium (Figure \ref{fig:right_k}). 

The right amount of flow rate $k$ is when the local slope reaches the angle of repose after a single step, which is the equilibrium state. If the perfect amount of flow to the adjacent block is $h_\Delta$, which is 1 unit in Figure~\ref{fig:right_k}, then to find the perfect $k$ from equation \ref{eq:erosion}, we need $k$ so that the amount of flow $\Delta h$ is the perfect amount $h_\Delta$,
\begin{eqnarray}\label{eq:perfect}
q &=& k\Delta x\left(\frac{h_2-h_1}{dx} - \frac{(h_2-h_\Delta) - (h_1 + h_\Delta)}{dx}\right) \nonumber\\
\Delta h &=& \frac{1}{\Delta x} k \Delta x \left(\frac{2h_\Delta}{dx}\right) \\
k &=& \frac{\Delta x}{2} \nonumber
\end{eqnarray} 
We observe the rate of flow is proportional to the grid size, and it is $k=\Delta x/2$ for the 1D case. For 2D with eight-point connectivity and using similar reasoning, $k=\Delta x^2/8$ works best, as validated experimentally. 

Another method to speed up the simulation is by updating only a portion of the height map. Because the soil erosion happens only near a tool interaction, we can have a bounding box of update so that we update only a portion of the height map. Instead of a bounding box we could have checked the grids to update only when change happens, however this method required more computation time.

\subsection{A* Planner Order of Complexity}
The problem of finding the optimal single stroke path for drawing actually grows rather quickly. For trenching the S shape, the planner had to open 788,492 nodes, although the shape looks simple. Every step after the planner has to back-track (for S, after the first U-turn), the number of nodes it has to open grows by $4^n$ order. Because a single step backwards is equivalent to growing the search tree, it increases by four-fold for our four connectivity problem. By changing the ratio $\alpha$, we can aggressively look for path that is suboptimal but that opens considerably fewer nodes. For the S example, by reducing alpha from 7.2 to 3.0, the planner had to open 63,568 nodes to find the same optimal path of 27 steps. So for this case it was still able to find the optimal path without opening as many nodes. In the T example, by reducing alpha from 7.2 to 3.0, the planner had to open 820 nodes to find path that took 21 steps (while the optimal path has 18 steps).

\section{Conclusions}
\label{sec:conclusion}
This work introduced a new interesting problem: shaping sand autonomously. Although it is baffling to think about how we can start tackling the problem, by dividing the tasks into simpler ones and by formulating the problem into something more concrete, we can find different ways to go about this. The simple simulator shows great promise since we can incorporate this with other sand simulation methods (like force interaction which is important in actual robots). The previous works we looked at would be a good starting point in the future. Another interesting formulation is the A* planner. We were able to simplify the problem into discrete form and actually found a interesting problem. A single stroke shaping is somewhat like the traveling salesman problem. We can play with the heuristics to find a faster algorithm that can find the optimal path without opening too many nodes. 

Future work will apply deep reinforcement learning for shaping sand. Because a deep neural network will be able to capture the high dimensionality of sand, it will be able to shape sand without the need for simplification of the sand model as it was necessary for A-star. Also, force-interaction can be added to the model which will allow more accurate representation of the sand and robot dynamics. We believe this work provides a good starting point for shaping continuously deformable environment for many situations. 


%
% ...

\bibliographystyle{asmems4}

\bibliography{asme2e}

\appendix       
\section*{Appendix A: Sand model comparison}

\begin{figure}[tb]
\centering
\includegraphics[width=0.48\textwidth]{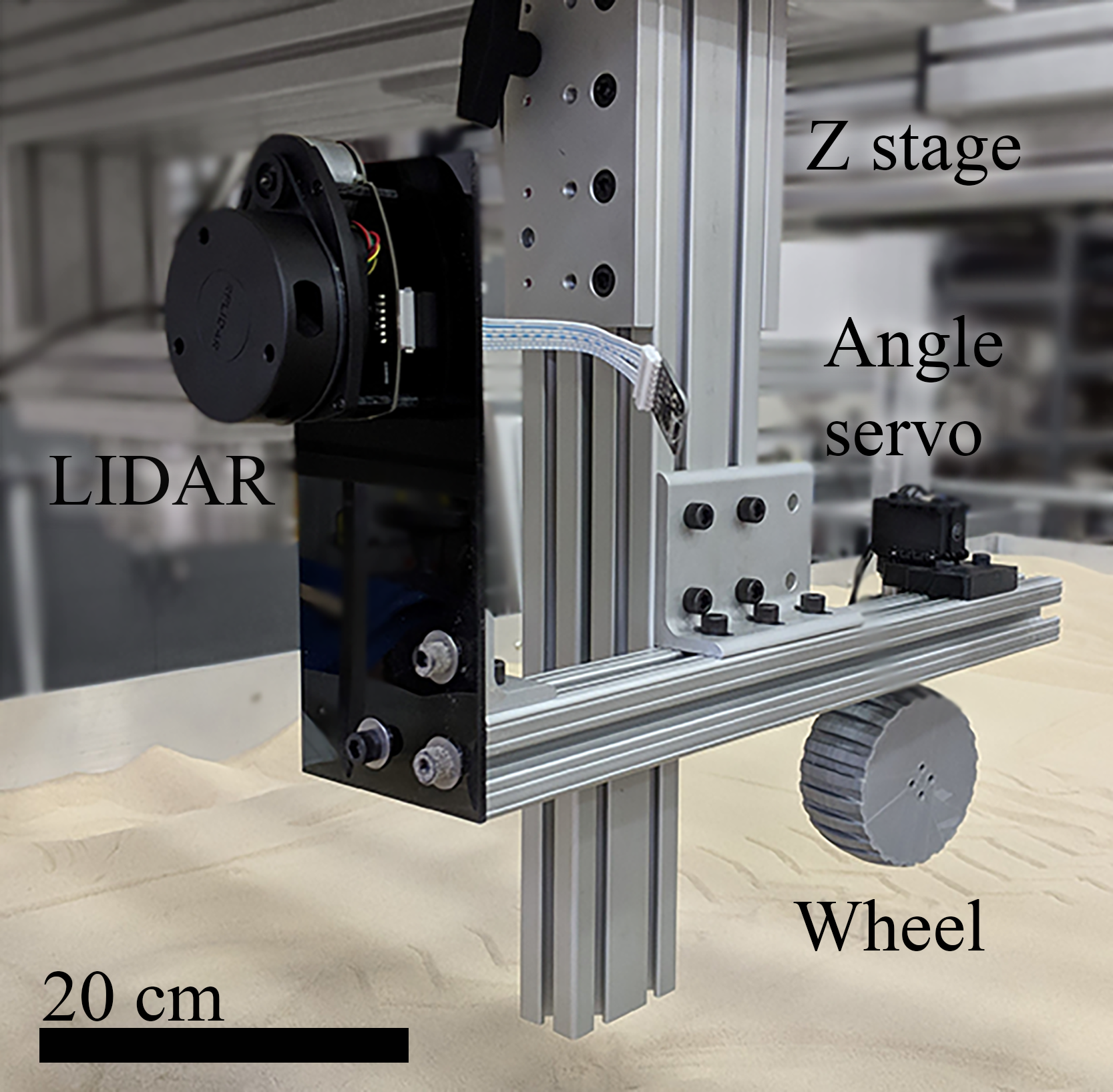}
\caption{The single-wheel testbed setup (right). For each trench, the wheel is lowered to a fixed sinkage ($h_0$) and moves across level sand at a fixed slip ratio ($s$) and slip angle ($\beta$). \cite{Pavlov}}
\label{fig:testbed}
\end{figure}

\begin{figure}[tb]
    \centering
    \includegraphics[width=0.48\textwidth]{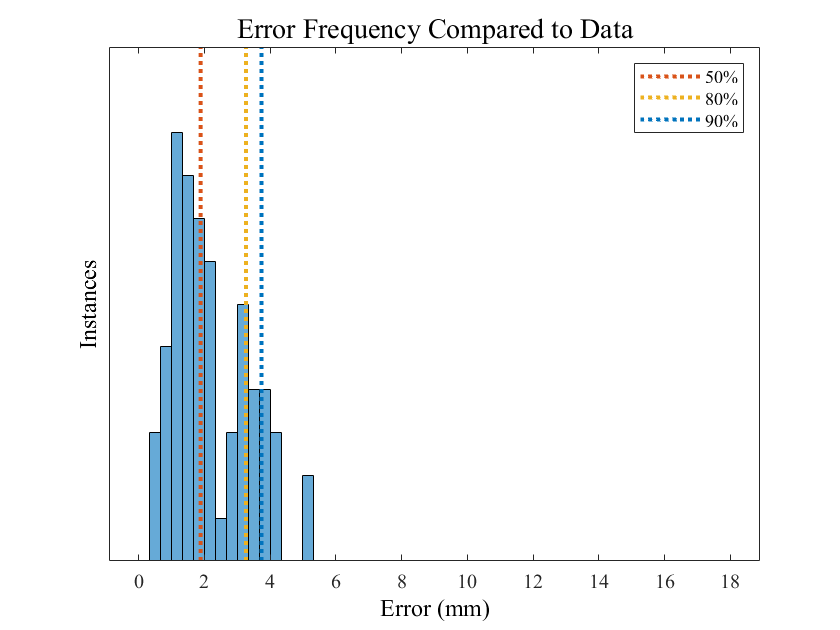}
    \caption{Frequency of errors in proposed model compared to trench data collected in \cite{Pavlov}.}
    \label{fig:bk_error}
\end{figure}

\begin{figure}[tb]
    \centering
    \includegraphics[width=0.48\textwidth]{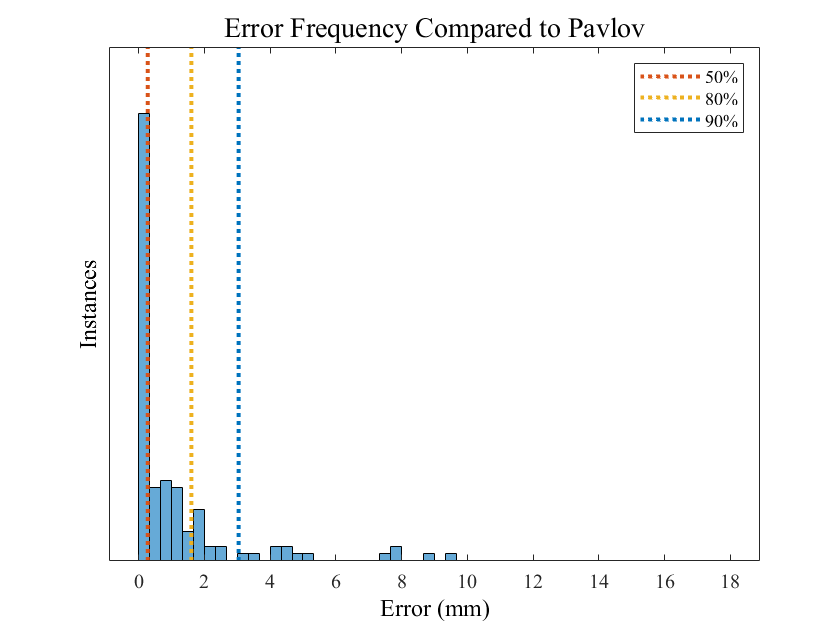}
    \caption{Frequency of errors in proposed model compared to trench model presented in \cite{Pavlov}.}
    \label{fig:bk_cp_error}
\end{figure}

Figure \ref{fig:testbed} shows the test bed we used to compare trench profiles \cite{Pavlov}, with system parameters listed in Table~\ref{table:trench_params}. The results are listed in Table~\ref{table:results_grouser}, and we show a histogram of height errors compared to the experimental data (Figure~\ref{fig:bk_error}) and the first principles model (Figure~\ref{fig:bk_cp_error}) reported in \cite{Pavlov}.

\begin{table}[tb]
\centering
  \begin{tabular}{ |c|c|c| }
  \hline
  Sym. & Description & Value\\
  \hline
  $\phi$ & Angle of repose [ $^{\circ}$ ] & 29\\
  $c$ & Cohesion [Pa] & 0\\
  \hline
  $r$ & Wheel radius [mm] & 48.0\\
  $b$ & Wheel width [mm] & 50.0\\
  $h_g$ & Grouser height [mm] & 5.0 \\
  $\zeta$ & Grouser volume fraction & 0.1\\
  \hline 
  \end{tabular}
   \caption{Soil and wheel parameters used for model validation.}
   \label{table:trench_params}
\end{table}

\begin{table}[tb]
\centering
    \setlength{\tabcolsep}{4.5pt}
  \begin{tabular}{ |c|c|c|c|c|c| }
  \hline
  Trench Type& Avg. Error& Median Error& Depth Error\\
  ~ & [mm] & [mm] & [mm]\\
  \hline
  \multicolumn{4}{|c|}{Comparison to Trench Data}\\
  \hline
    All trenches & 2.2 & 1.7 & 0.9\\
    \hline
    $\beta = 0^\circ$ & 2.3 & 1.4 & 1.2\\
    $\beta = 22.5^\circ$ & 2.0 & 1.7 & 0.5\\
    $\beta = 45^\circ$ & 2.1 & 1.6 & 0.7\\
    $\beta = 67.5^\circ$ & 2.1 & 1.4 & 0.2\\
    $\beta = 90^\circ$ & 2.6 & 2.0 & 1.7\\
    \hline
    $h_0 = 5$mm & 2.0 & 1.6 & 0.7\\
    $h_0 = 15$mm & 2.1 & 1.6 & 0.8\\
    $h_0 = 25$mm & 2.6 & 2.6 & 1.1\\
    \hline 
\multicolumn{4}{|c|}{Comparison to \cite{Pavlov}}\\
  \hline
    All trenches & 1.1 & 0.4 & 0.6\\
    \hline
    $\beta = 0^\circ$ & 0.5 & 0.4 & 0.03\\
    $\beta = 22.5^\circ$ & 1.6 & 0.4 & 0.3\\
    $\beta = 45^\circ$ & 1.7 & 0.7 & 0.7\\
    $\beta = 67.5^\circ$ & 1.2 & 0.5 & 0.6\\
    $\beta = 90^\circ$ & 0.4 & 0.1 & 0.2\\
    \hline
    $h_0 = 5$mm & 0.4 & 0.1 & 0.1\\
    $h_0 = 15$mm & 0.6 & 0.4 & 0.1\\
    $h_0 = 25$mm & 2.2 & 1.9 & 0.9\\
    \hline 
  \end{tabular}
  \caption{Comparison of proposed model to trench measurements and model proposed in \cite{Pavlov}}
  \label{table:results_grouser}
\end{table}



\end{document}